\documentclass[letterpaper, 10 pt, journal, twoside]{IEEEtran}
\usepackage{times}
\usepackage{graphicx}
\usepackage{amsmath}
\usepackage{amssymb}
\usepackage{multirow}
\usepackage{booktabs}
\usepackage{fancyhdr}

\begin{document}

\title{Hierarchical Point Cloud Encoding and Decoding with Lightweight Self-Attention based Model}

\author{En Yen Puang$^{1}$, Hao Zhang$^{2}$, Hongyuan Zhu$^{1}$~\IEEEmembership{Member,~IEEE,} and Wei Jing$^{1,3}$%
\thanks{Manuscript Received September 9 2021; Revised December 30 2021; Accepted January 25 2022.}%
\thanks{This paper was recommended for publication by Editor Cesar C. Lerma upon evaluation of the Associate Editor and Reviewers’ comments. This work was supported by AME Programmatic Funding Scheme (Project \#A18A2b0046), Career Development Fund (Project C210812033) and RobotHTPO Seed Fund (Project C211518008).}%
\thanks{$^{1}$E. Y. Puang, H. Zhu and W. Jing are with Institute for Infocomm Research, A*STAR, Singapore {\tt\footnotesize  \{puang\_en\_yen, zhuh\}@i2r.a-star.edu.sg}}%
\thanks{$^{2} $H. Zhang is with Centre for Frontier AI Research, A*STAR, Singapore {\tt\footnotesize zhang\_hao@ihpc.a-star.edu.sg}}%
\thanks{$^{3} $W. Jing is also with Alibaba Group, China {\tt\footnotesize 21wjing@gmail.com}}%
\thanks{Digital Object Identifier (DOI): 10.1109/LRA.2022.3149569}
}

\markboth{IEEE Robotics and Automation Letters. Preprint Version. Accepted January, 2022}{Puang \MakeLowercase{\textit{et al.}}: Hierarchical Point Cloud Encoding and Decoding with Lightweight Self-Attention based Model} 

\maketitle

\thispagestyle{fancy}
\lhead{\scriptsize IEEE ROBOTICS AND AUTOMATION LETTERS. PREPRINT VERSION. ACCEPTED JANUARY, 2022}
\rhead{\scriptsize \thepage}
\renewcommand{\headrulewidth}{0pt}
\fancyfoot{}
\cfoot{\scriptsize{\copyright~2022 IEEE. Personal use of this material is permitted.  Permission from IEEE must be obtained for all other uses, in any current or future media, including reprinting/republishing this material for advertising or promotional purposes, creating new collective works, for resale or redistribution to servers or lists, or reuse of any copyrighted component of this work in other works.}}

\begin{abstract}
In this paper we present SA-CNN, a hierarchical and lightweight self-attention based encoding and decoding architecture for representation learning of point cloud data. 
The proposed SA-CNN introduces convolution and transposed convolution stacks to capture and generate contextual information among unordered 3D points. 
Following conventional hierarchical pipeline, the encoding process extracts feature in local-to-global manner, while the decoding process generates feature and point cloud in coarse-to-fine, multi-resolution stages.
We demonstrate that SA-CNN is capable of a wide range of applications, namely classification, part segmentation, reconstruction, shape retrieval, and unsupervised classification. While achieving the state-of-the-art or comparable performance in the benchmarks, SA-CNN maintains its model complexity several order of magnitude lower than the others. In term of qualitative results, we visualize the multi-stage point cloud reconstructions and latent walks on rigid objects as well as deformable non-rigid human and robot models.
\end{abstract}

\begin{IEEEkeywords}
Visual Learning, Deep Learning for Visual Perception, Recognition
\end{IEEEkeywords}

\section{Introduction} \label{sec:intro}

\IEEEPARstart{W}{ith} recent advancement on 3D sensory devices and the surging needs on 3D perception in applications such as the robotics and autonomous vehicles \cite{potapova2017survey}, understanding point cloud data from 3D sensors becomes increasingly important. Unlike 2D pixel-based images, 3D point cloud data is unordered and unstructured. Moreover, point clouds are usually in large scale and thus effective representation learning is often desirable for many downstream tasks \cite{achlioptas2018learning}.

Many deep learning based approaches \cite{guo2020deep} have been applied to 3D point cloud perception. Early work rely on handcrafted 3D features which require intensive feature engineering and are difficult to generalize. 
Then methods like \cite{maturana2015voxnet, riegler2017octnet} convert 3D point cloud into voxel representation and extend 2D convolution to 3D, but bear high computational cost and low voxel resolution. 
Some work \cite{su2015multi, yang2019learning} explore multi-view projection methods that apply 2D CNN to extract the view-wise features circumvent the heavy computational complexity, but usually requires heuristics in choosing informative viewpoints. 
Later, \cite{qi2017pointnet, qi2017pointnet++} propose an efficient and effective end-to-end architecture that directly learns the representations from the unstructured point cloud data. 

\begin{figure}[t]
  \begin{center}
    \includegraphics[width=0.65\linewidth]{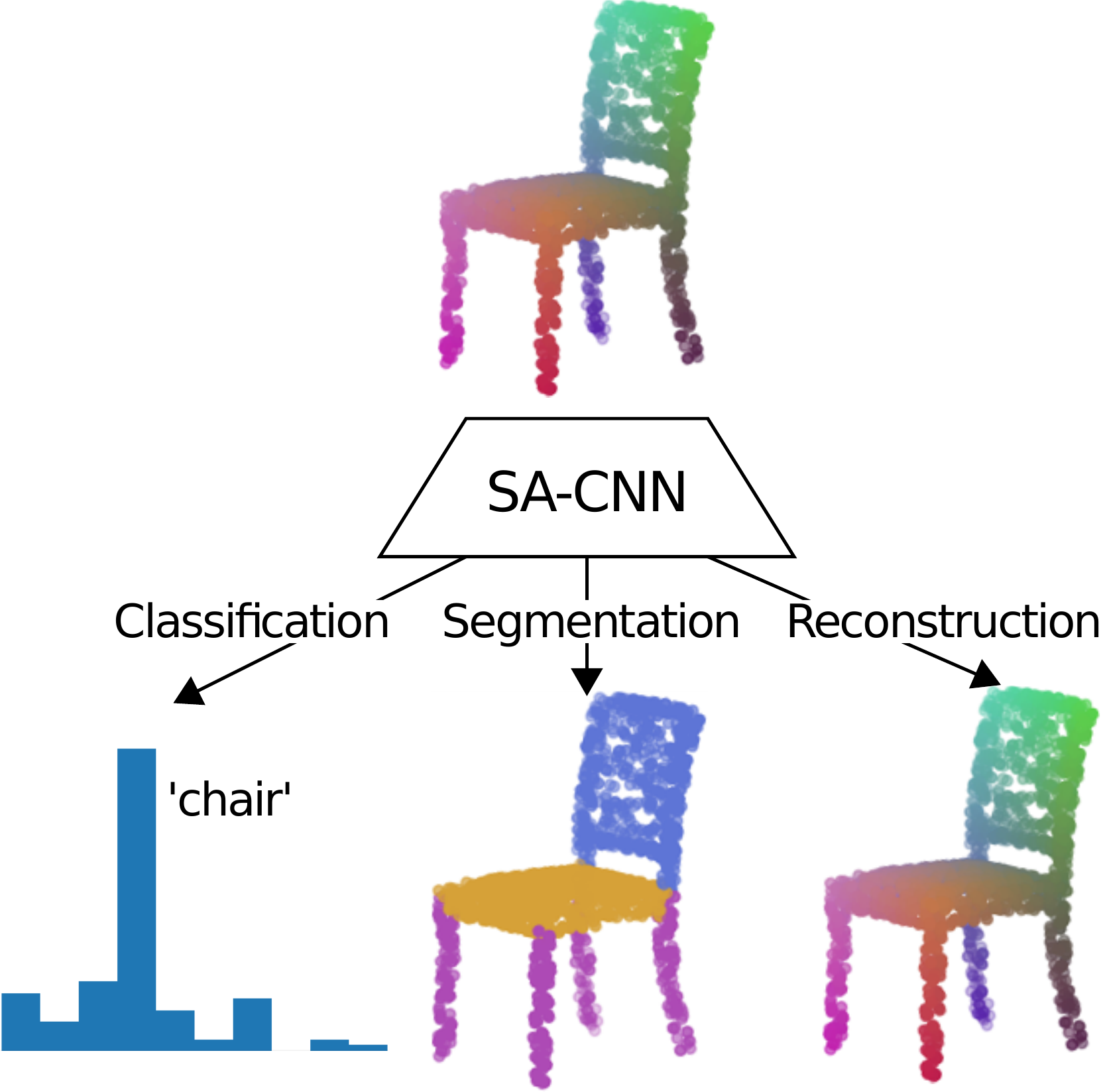}
  \end{center}
  \caption{SA-CNN is a lightweight self-attention based architecture capable for point cloud analysis tasks such as classification, segmentation and reconstruction by auto-encoding.}
\label{fig:intro}
\end{figure}

In term of dimensionality reduction, auto-encoder or encoder-decoder architecture has been widely adopted in compressing and/or distilling input into lower dimensional representation in multiple domains. 
In the context of point cloud, earlier methods \cite{qi2017pointnet, groueix1802atlasnet, achlioptas2018learning} use mostly MLP in feature extraction and generation. Hierarchical pipeline was later introduced \cite{qi2017pointnet++, yang2018foldingnet, shu20193d} to encode/extract feature in local-to-global manner, and to decode/generate point cloud in coarse-to-fine, multi-resolution stages.

To this end, we propose to apply a lightweight hierarchical self-attention mechanism \cite{vaswani2017attention, qi2017pointnet++} to handle the unstructured, interrelated contextual information of point cloud data. Self-attention models have been widely adopted in natural language processing, vision and point cloud \cite{chen2019gapnet, feng2020point} due to their efficiency in handling dependencies in sequential/unordered data. 
Leveraging on the efficiency and simplicity of the self-attention mechanism, we propose a lightweight point cloud convolution and transposed convolution operator for hierarchical feature extraction and generation on point cloud data. We demonstrate that our models are capable for a wide range of applications and achieve the state-of-the-art or comparable performance while keeping the model complexity low.
Hence, the main contributions of this work are summarized as follows:
\begin{itemize}
    \item we propose a hierarchical network architecture named SA-CNN for point cloud encoding and decoding using an lightweight self-attention based model;
    \item we demonstrate the applications of SA-CNN in classification, part segmentation, reconstruction, shape retrieval and unsupervised classification;
    \item we show that SA-CNN achieves the state-of-the-art or comparable performances in multiple benchmarks while keeping the model complexity orders of magnitude lower.
\end{itemize}

\section{Related Work} \label{sec:literature}
\subsection{Point-based Methods for Point Cloud Analysis}
Point-based methods became popular due to its efficiency, flexibility and scalability. Different network architectures e.g., Multi-Layer Perceptron (MLP), point convolution, graph-based, and attention-based network could be used to deal with the point-wise feature extraction. \cite{qi2017pointnet,qi2017pointnet++} propose PointNet(++) as pioneer work for point-based methods on 3D point cloud data, which directly learn feature representation from the point cloud data.
Convolution-based methods have also been applied to point cloud data~\cite{li2018pointcnn,thomas2019kpconv} by directly performing convolution-like operation on 3D points.
More recently, graph-based neural networks~\cite{wang2019dynamic,te2018rgcnn} and attention-based methods~\cite{zhao2021point, guo2020pct, yang2020attpnet} also demonstrate good performance for point cloud feature learning. However, these related works only learn features in encoding manner. Our work is a point-based method that uses self-attention mechanism to build hierarchical encoding and decoding models for efficient representation learning of point cloud.

\subsection{Self-Attention Mechanism}
Self-attention mechanism has become an integral part of various neural network-based models for handling the sequence inputs, allowing modeling of dependencies without regard to their distance in the sequences. 
Specifically, the Transformers \cite{vaswani2017attention} have revolutionized natural language processing tasks due to their tremendous ability on representation learning. 
Recently, there are several work \cite{chen2019gapnet,feng2020point} adopt the attention mechanism to address the 3D point cloud problems. 
\cite{yang2019modeling,engel2021point,yan2020pointasnl} proposed additional mechanism together with self-attention to improve point sampling and feature aggregation.
These self-attention based methods have achieved promising performance on various point cloud tasks, but their models are still less efficient with a large amount of parameters and high complexity. Our methods adopt the same self-attention mechanism in the design of a lightweight architecture to achieve the state-of-the-art or comparable performance with orders of magnitude lesser model parameter and computational complexity. 

\subsection{Point Cloud Auto-Encoding}
Point cloud generation are generally divided into 2 catogories: Generative Adversarial Network (GAN) based and non-GAN based methods. GAN-based generative model \cite{achlioptas2018learning, li2018point, shu20193d, ramasinghe2020spectral} sample point cloud based on the data distribution learned during training. These methods often output realistic point cloud with great details, but generally require more complex training procedures. 

Non-GAN based methods are mostly in encoder-decoder, unsupervised architecture \cite{fan2017point}. To improve reconstruction performance, \cite{gadelha2018multiresolution} requires the point cloud to be a 1D ordered list structure. \cite{yang2018foldingnet, groueix1802atlasnet} parameterize the decoder with 2D grid patches as part of the latent representation, while \cite{ramasinghe2020spectral} parameterizes point cloud with spectral frequencies. Our auto-encoder is a non-GAN, self-supervised based architecture that reconstructs input point cloud based only on a latent vector and without any other parameterization.

\section{Preliminary}
\subsection{Self-Attention Layer}
The self-attention layer $ \mathcal{A}(\cdot)$ we adopt is a scaled dot-product attention mechanism \cite{vaswani2017attention} $\mathcal{A} : Q, K, V \to \mathbb{R}^{N \times d'}$ which takes as input \textit{Query} $Q \in \mathbb{R}^{N \times d}$, \textit{Key} $K \in \mathbb{R}^{N' \times d}$ and \textit{Value} $V \in \mathbb{R}^{N' \times d'}$, and aggregates information from $V$ for each token in $Q$ based on the alignment with $K$: 
\begin{align}
    \label{eqn:selfattention}
    \mathcal{A}(Q,K,V) = \sigma \left ( \frac{QK^\intercal}{\sqrt{d}} \right )V 
\end{align}
where $\sigma$ is a softmax operator. Multi-head attention simply concatenates the output of multiple self-attention layer. The output is of $\mathbb{R}^{N \times hd^\prime}$ where $h$ is the number of head.

Point cloud data is here defined as an unordered set of $N$ points $\mathcal{P} = \big \{ \boldsymbol{p}_i \big \}^N_{i=1} \, \big \vert \, \boldsymbol{p} \in \mathbb{R}^3$ in 3D Euclidean space without additional features.
When applied on point cloud processing, self-attention layer acts analogously to a convolution kernel. \textit{Query} points are analogous to the kernel's center. Whereas the neighbouring points, in which the \textit{Key} and \textit{Value} are derived from, are analogous to the rest of the kernel inputs. Multi-head attention is then analogous to a convolution filter.
Self-attention model is well suited in handling point cloud \cite{chen2019gapnet, feng2020point, zhang2020attan, wang2019exploiting} because: 
\begin{itemize}
    \item Self-attention is invariant to input permutation and hence able to handle input point cloud data in the form of unordered set.
    \item Instead of one-to-many relationship rendered by convolution kernel, self-attention improves points interactions by having many-to-many relationship in the input set in an efficient manner.
    \item With adequate normalization in the input set, self-attention is robust under rigid transformation such as translation and rotation.
\end{itemize}

\subsection{Processing Pipeline}
Unlike CNN's grid structure, point cloud requires several pre-processing layers in a hierarchical pipeline \cite{qi2017pointnet++}:
\begin{itemize}
    \item \textit{Sampling}. Given a set of points, a subset is selected and formed into an unstructured probing grid.
    \item \textit{Grouping}. Given a set of points and its probing grid, the nearest neighbours are gathered into local neighbourhoods for each probing points.
    \item \textit{Normalizing}. Given a set of local neighbourhoods, points are normalized by subtracting its probing point and scaling by a constant.
    \item \textit{Interpolating}. Given neighbourhoods from 2 adjacent levels, features are propagated from the smaller neighbourhood with distance based interpolation. 
\end{itemize}

\section{Model Architecture} \label{sec:method}
The proposed SA-CNN consists of 2 basic operators for processing data in the form of unordered set. We will first discuss these operators, then we describe how to build with these operators for different applications.

\subsection{SA-CNN Modules}

\begin{figure}[t]
    \centering
    \includegraphics[width=0.7\linewidth]{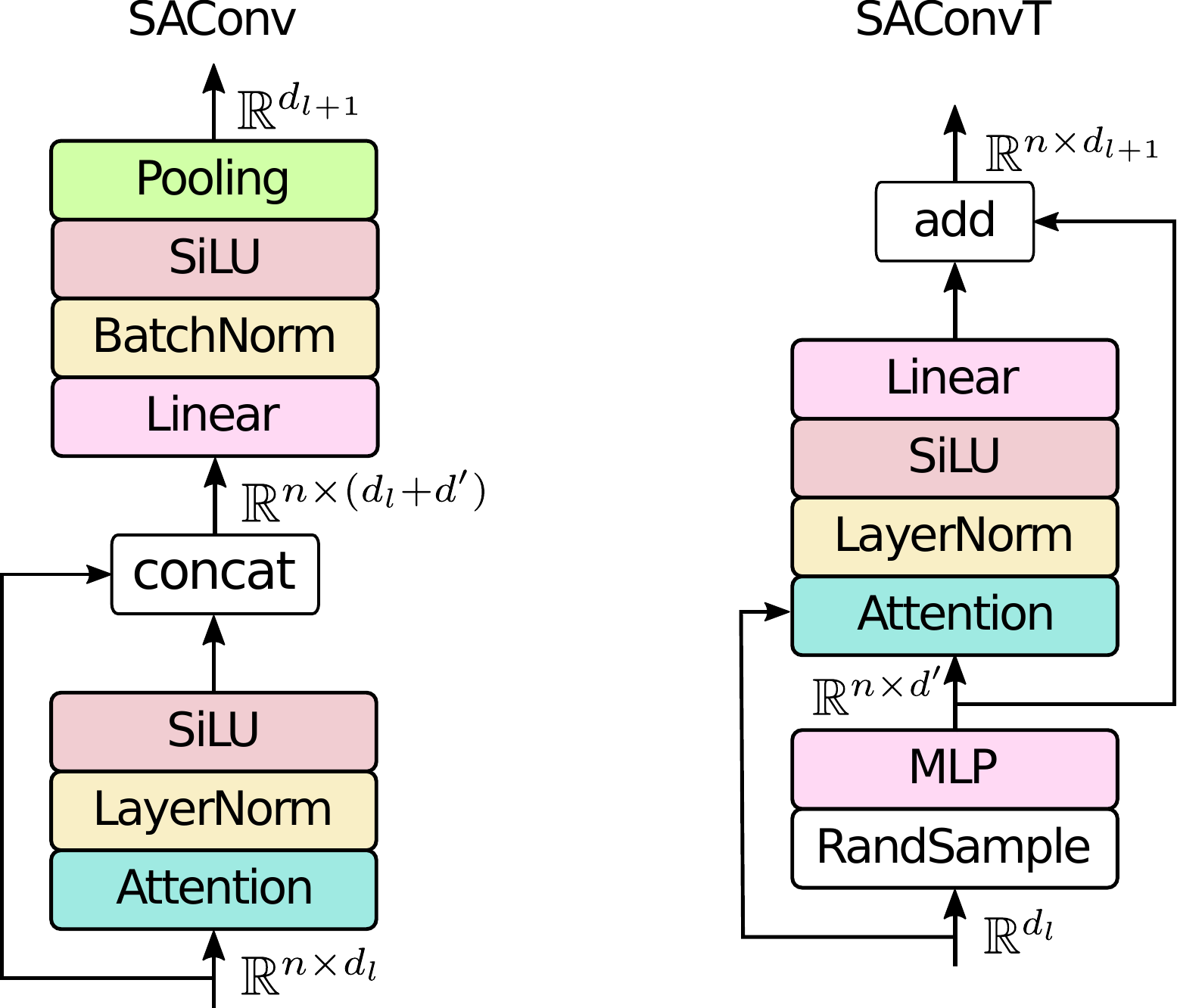}
   \caption{The design of SAConv and SAConvT stacks. The former aggregates features from a set, while the latter expands a feature into a set.}
\label{fig:saconv}
\end{figure}

\textbf{SAConv} is a set convolution stack that aggregates features $\mathrm{SAConv}: \mathbb{R}^{n \times d_l} \to \mathbb{R}^{d_{l+1}}$ from an unordered set of size $n$ and outputs a single feature vector. 
For a self-attention layer to work analogous to a convolution kernel on point cloud, \textit{Sampling-Grouping-Normalizing} (SGN) pre-processing layer is deployed to form the input $\mathcal{X} \in \mathbb{R}^{(k+1) \times d}$ where $d$ is the feature dimension and $k$ is the number of nearest neighbour of a probing point. \textit{Query}, \textit{Key} and \textit{Value} in Eq \ref{eqn:selfattention} are from
\begin{subequations}
\begin{align}
    \label{eqn:qkv}
    Q &= \mathcal{X}W_q \\
    K &= \mathcal{X}W_k \\
    V &= \mathcal{X}W_v
\end{align}
\end{subequations}
the linear layers where $W_{q,k} \in \mathbb{R}^{d \times d_q}$ and $W_v \in \mathbb{R}^{d \times d'}$ are the weights. The output of the multi-head attention then goes through a sequence of common layers as depicted in Fig. \ref{fig:saconv} (left). A pooling layer is placed at the end as the symmetric function that aggregates the final features in the set. 

\textbf{SAConvT} is a transposed convolution stack proposed to generate an unordered set $\mathrm{SAConvT}: \mathbb{R}^{d_l} \to \mathbb{R}^{n \times d_{l+1}}$ based on a single input feature vector $\mathcal{X}$. It consists of a random sampling layer that draw $\mathcal{N} \in \mathbb{R}^{n \times d'}$ samples from unit Gaussian distribution and
\begin{align}
    \label{eqn:convt}
    \mathcal{X}' = \mathrm{MLP} \Big (\mathrm{concat}_d \big (\mathcal{X}, \, \mathcal{X}W_r \ast \mathcal{N} \big) \Big )
\end{align}
where $W_r \in \mathbb{R}^{d \times d'}$ is the weight of the linear layer before the broadcasted element-wise multiplication $\ast$ with $\mathcal{N}$, and $\mathrm{concat}_d$ is concatenation along feature dimension. Multi-head attention takes it as input $\mathcal{A}(\mathcal{X}')$ and then followed by a sequence of common layers as depicted in Fig~\ref{fig:saconv} (right).

\textbf{Positional Encoding} were proposed \cite{vaswani2017attention} to provide sequential information for each tokens. Although unordered, points in point cloud are well represented in euclidean space. Therefore, we use the 3D \textit{xyz} coordinate of point as its positional encoding. Combining \textit{Normalizing} pre-processing layer, the positional encoding is its normalized coordinate with respect to the probing point of the neighbourhood. In the hierarchical pipeline, $d_l=3+d_{l-1}$ given $d_0 = 0$ indicates the input dimension of level $l$. Concretely, in each hierarchical levels the input to SAConv is the concatenation of the normalized 3D coordinate and the features computed from the previous level given the input dimension is 3.

\subsection{Applications of SA-CNN}

\begin{figure}[t]
    \centering
    \includegraphics[width=0.9\linewidth]{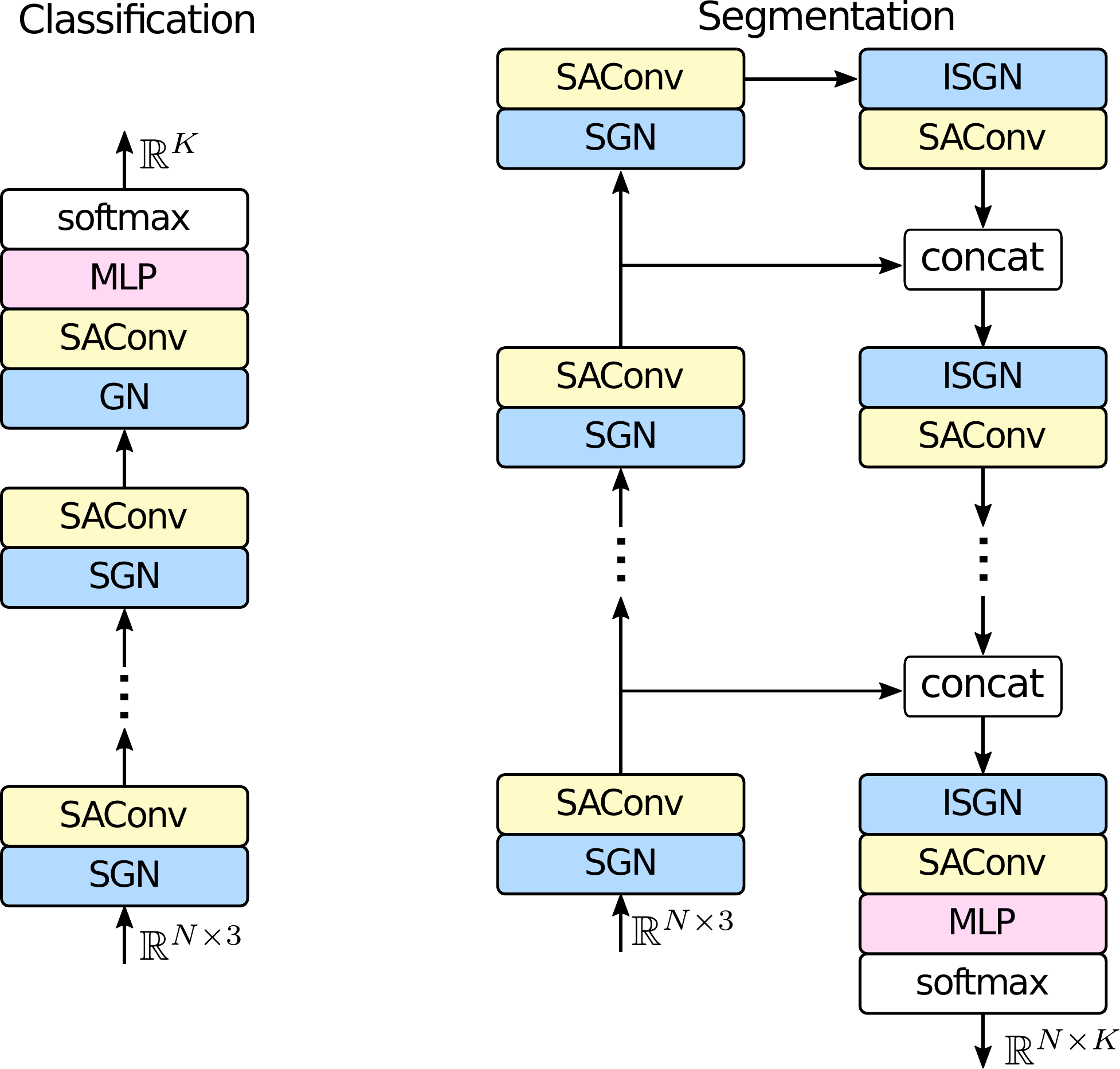}
  \caption{The architectures of classification and segmentation networks. Both networks use SAConv paired with pre-processing layers \cite{qi2017pointnet++} (I/SG/N) in each hierarchical levels.}
\label{fig:class_segm}
\end{figure}

\subsubsection{Classification and Part Segmentation}
We apply SA-CNN modules as a drop-in replacement to solve the point cloud classification and part segmentation tasks using the hierarchical feature extraction pipeline \cite{qi2017pointnet++} as depicted in Fig \ref{fig:class_segm}. Classification network is composed of several levels, each consists of a SGN-SAConv pair. Each level abstracts a subset of probing points, while the last level groups (GN, without \textit{sampling}) all the remaining for the final classification output. The part segmentation network adopts U-Net \cite{ronneberger2015unet} structure with skip connections. ISGN combines \textit{interpolating} layer with SGN and is used to propagate features to the next level. 

\subsubsection{Auto-encoding}
\begin{figure}[t]
    \centering
    \includegraphics[width=0.85\linewidth]{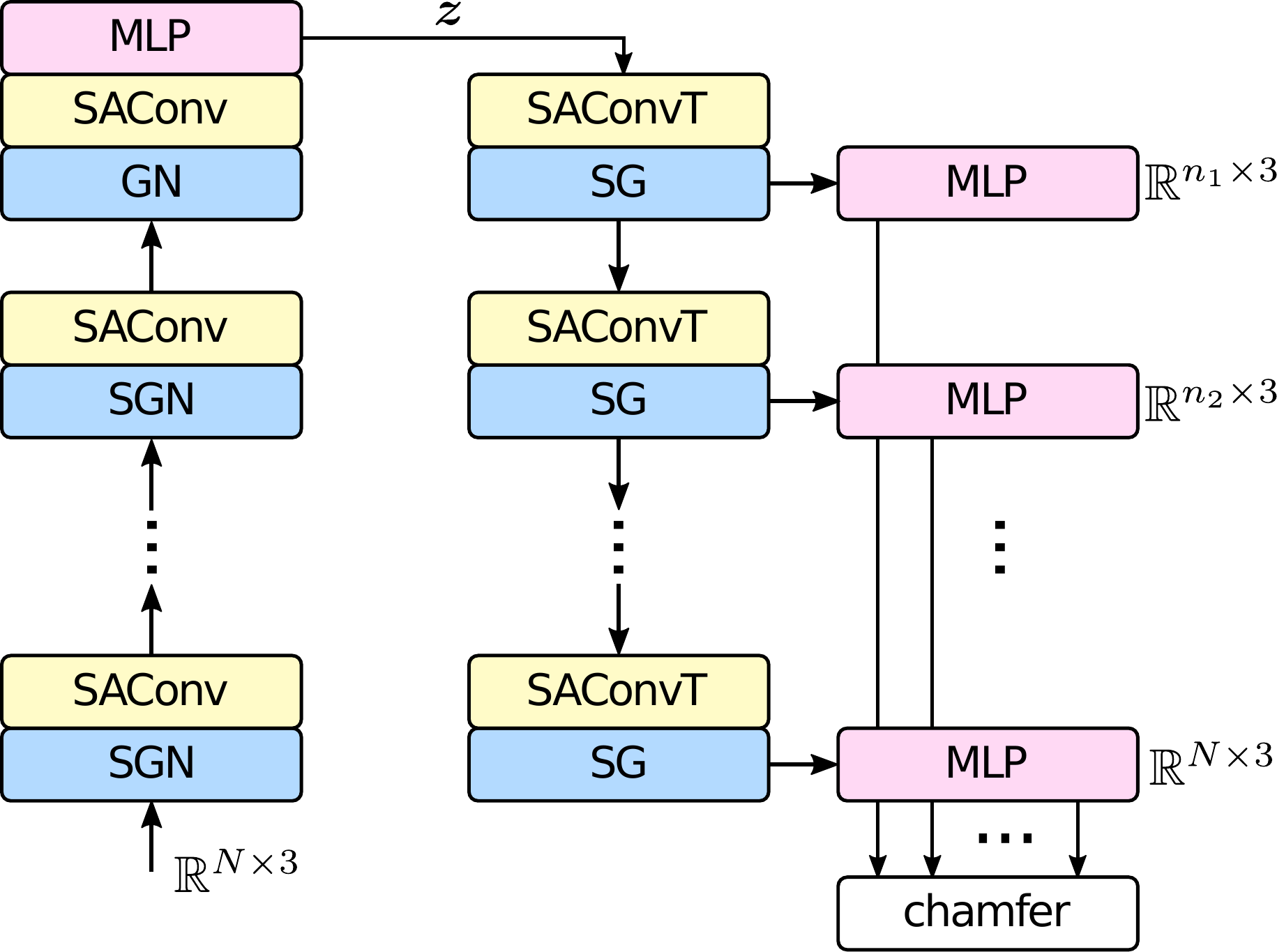}
  \caption{The architecture of auto-encoder network uses SAConvT for expanding latent representation $\boldsymbol{z}$ into points. Each level in the decoder outputs a set/subset of points and contributes to the final chamfer distance.}
\label{fig:autoencoder}
\end{figure}

We apply SA-CNN modules to solve the point cloud reconstruction task with a hierarchical auto-encoder architecture depicted in Fig \ref{fig:autoencoder}. Similar to the pipeline of the classification task, the \textbf{encoder} $f: \mathcal{P} \mapsto \boldsymbol{z}$ takes in as input a set of points $\mathcal{P} \in \mathbb{R}^{N \times 3}$ and processes it with several SGN-SAConv pairs. A MLP then turns the features into a latent representation vector $\boldsymbol{z} \in \mathbb{R}^L$ where $L \ll N$. 

The hierarchical \textbf{decoder} $g: \boldsymbol{z} \mapsto \big \{ \bar{\mathcal{P}_l} \big \}_{l=1}^H$ takes as input the latent vector, and expands it through $H$ levels of SG-SAConvT pairs. Each hierarchical level expands every points in the level and produces a set of output points $\bar{\mathcal{P}}$ given $|\bar{\mathcal{P}}_a| < |\bar{\mathcal{P}}_b|$ if $a<b$ and $|\bar{\mathcal{P}}_H|=N$.

\begin{figure}[t]
    \centering
    \includegraphics[width=0.95\linewidth]{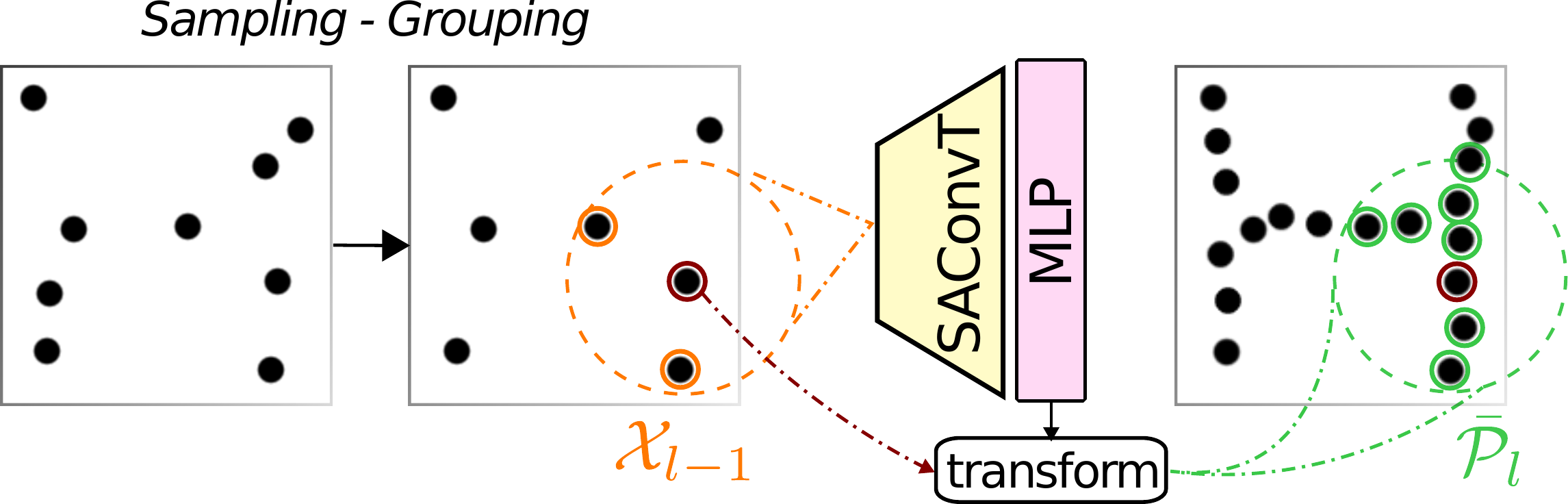}
  \caption{Points decoding process by SAConvT not only involves $n_i$ random samples but also $k$ neighbours in the preceding level. This added extra $\textit{key}$ and $\textit{value}$ to the self-attention layer in SAConvT stack. }
\label{fig:decoder}
\end{figure}

The multi-head attention in the SAConvT stack of the decoder takes as input a point feature $\mathcal{X}_l$ as well as  the feature of its nearest neighbours in the previous level $\bar{\mathcal{P}}_{l-1}$. After Eq \ref{eqn:convt} the \textit{Query}, \textit{Key} and \textit{Value}:
\begin{subequations}
\begin{align}
    \label{eqn:qkv_ae}
    Q &= \mathcal{X}'_lW_q \\
    K &= \mathrm{concat}_n \big ( \mathcal{X}'_lW_k, \, \mathcal{X}_{l-1}W_{k'} \big ) \\
    V &= \mathrm{concat}_n \big ( \mathcal{X}'_lW_v, \, \mathcal{X}_{l-1}W_{v'} \big )
\end{align}
\end{subequations}
where $W_{q,k,v,k',v'}$ are the weights of the linear layers, and $\mathrm{concat}_n$ is concatenation along token dimension. A MLP then turns the features of the level into points in euclidean space by scaling and translating it to the input point's coordinate frame. Fig \ref{fig:decoder} depicts the decoding process. 

The objective of the auto-encoder is to reconstruct the input point cloud by minimizing chamfer distance \cite{achlioptas2018learning}
\begin{equation}
\begin{split}
    \label{eqn:chamfer}
    \mathrm{chamfer}(\mathcal{P}_1, \mathcal{P}_2) =& \frac{1}{|\mathcal{P}_1|}\sum_{p \in \mathcal{P}_1} \min_{q \in \mathcal{P}_2} \| p - q \|^2_2  + \\
    &\frac{1}{|\mathcal{P}_2|}\sum_{q \in \mathcal{P}_2} \min_{p \in \mathcal{P}_1} \| p - q \|^2_2
\end{split}
\end{equation}
between the input points $\mathcal{P}$ and the output point $\bar{\mathcal{P}}_l$ 
\begin{align}
    \label{eqn:recon}
    \mathcal{L}_{rcs} = \sum_{l=1}^{H} \alpha_l \: \mathrm{chamfer}(\mathcal{P}, \bar{\mathcal{P}}_l)
\end{align}
in each levels where $\alpha_l \propto |\bar{\mathcal{P}}_l|$ and $\alpha_H = 1$. Hence, the reconstruction loss $\mathcal{L}_{rcs}$ is the weighted summation of the chamfer distances from all levels. This improves gradient propagation during the training of the auto-encoder. 

\subsubsection{Data Retrieval}
For the data retrieval task, \cite{liu2019densepoint} proposed to use the outputs of the penultimate fully-connected layer in the classification network as low dimensional representation of the input point cloud. In contrast, latent representation of point cloud trained with an auto-encoder can be used for the data retrieval task. Given an object point cloud, its latent representation should be closer to those from the same class in term of cosine distance. We proposed to use this latent representation for the point cloud shape retrieval task. To utilize labeled data, we include a triplet loss $\mathcal{L}_{tri}$ to better model the latent space during training:
\begin{align}
    \label{eqn:triplet}
    \mathcal{L}_{tri} = max \big (\textsc{cos}(\boldsymbol{z}, \boldsymbol{z}_n) - \textsc{cos}(\boldsymbol{z}, \boldsymbol{z}_p) + \alpha , 0 \big )
\end{align}
where $\textsc{cos}(\cdot)$ is the cosine similarity score ranging from 0 to 1, $\boldsymbol{z}_p$ and $\boldsymbol{z}_n$ are positive and negative inputs from the same and different class respectively.

\subsection{Implementation Details}
There are 4 hierarchical levels in the classification network. The pooling function used in SAConv stacks are [\textrm{Max}, \textrm{Max}, \textrm{Max}, \textrm{Avg}]. For segmentation network, there are $4 \times 2$ levels in the U-Net structure. Max Pooling is only used for the first 3 levels and Average Pooling for the rest. Category one-hot vector is concatenated to the feature before the last level. For auto-encoding network, there are $3 \times 2$ levels in the U-Net structure. For simplicity, $d_q = d_k = d_v$ sizes for \textit{Query}, \textit{Key} and \textit{Value} are the same in a self-attention layer. 

We use Farthest Point Sampling (FPS) and K-Nearest Neighbour (KNN) in \textit{Sampling} and \textit{Grouping} layers respectively. In \textit{Normalizing} layer we subtract probing point's coordinates from the neighbourhood and multiply different constants in each levels. In \textit{Interpolating} layer we use $exp(-\beta d)$ where $\beta = 16$ for weighted average of features based on point's relative distance $d$. \textrm{MLP} consists of multiple set of [\textrm{Dropout}, \textrm{Linear}, \textrm{BatchNorm}, \textrm{GELU}] sequence. 

\section{Experiment} \label{sec:exp}
In this section, we evaluate the proposed SA-CNN models in a series of comprehensive experiments. Concretely, we first evaluate our model in classification and part segmentation benchmarks. Then, demonstrate the ability in shape retrieval  and performance in unsupervised classification of our auto-encoder. Lastly, we visualize the results of point cloud reconstruction and illustrate the continuity of the latent space through latent walk in multiple object domains. 

\subsection{Shape Classification}
We evaluate SA-CNN on 3D point cloud classification task using ModelNet40 \cite{wu20153d} which consists of 9843 training models and 2468 test models in 40 classes. We uniformly sample 1024 points, zero the bounding box's center and normalize to unit sphere. During training we augment data with per-point Gaussian noise with std 0.001, random rotation between [-25, 25] degree in the vertical axis and [-5, 5] for the other 2 axes, scaling between [0.6667, 1.5] and translation between [-0.2, 0.2] in all 3 axes. During testing, we adopt voting strategy where 10 tests are performed with the same data augmentations except rotations and noise. 

\begin{table}[t]
    \small
    \centering
    \caption{ModelNet40 classification benchmark. Input consists of number of points, 'p'(point) and/or 'n'(normal).}
    \vspace{1mm}
	\begin{tabular}{llcc}
		\toprule
		Methods & Input & \#param(M) & Accuracy($\%$) \\
        \midrule
        PointNet~\cite{qi2017pointnet}        & 1k p & 3.5 & 89.2 \\
        PointNet++~\cite{qi2017pointnet++}    & 1k p & 1.5 & 90.7 \\
        PointNet++~\cite{qi2017pointnet++}    & 5k p,n & 1.5 & 91.9 \\
        Kd-Net~\cite{klokov2017escape}        & 32k p & 2.0 & 91.8 \\
        PointCNN~\cite{li2018pointcnn}        & 1k p & 0.3 & 92.5 \\
        DGCNN~\cite{wang2019dynamic}          & 1k p & 1.8 & 92.9 \\
        PCNN~\cite{atzmon2018point}           & 1k p & 1.4 & 92.3 \\
        DensePoint\cite{liu2019densepoint}    & 1k p & 0.7 & 93.2 \\
        RS-CNN\cite{liu2019relation}          & 1k p & 1.4 & 93.6 \\
        KPConv~\cite{thomas2019kpconv}        & 7k p & 14.3 & 92.9 \\
        Pnt Tfmer~\cite{engel2021point}       & 1k p,n & 13.50 & 92.8 \\
        Pnt Tfmer~\cite{zhao2021point}        & 1k p & 9.6 & \textbf{93.7} \\
        PCT~\cite{guo2020pct}                 & 1k p & 2.9 & 93.2 \\
        PointASNL~\cite{yan2020pointasnl}     & 1k p & - & 92.9 \\
        AttPNet \cite{yang2020attpnet}        & 1k p & - & 93.6 \\
        PAT \cite{yang2019modeling}           & 1k p & - & 91.7 \\

        \midrule
        SA-CNN (Ours)                         & 1k p & \textbf{0.04} & 92.3 \\
        \bottomrule
	\end{tabular}
	\label{tab:classification}
\end{table}

\begin{table}[t]
    \small
    \centering
    \caption{Computational complexity for ModelNet40 classification. All methods take as input 1024 points on a GeForce RTX 3090.}
    \setlength{\tabcolsep}{3pt}
    \begin{tabular}{c|ccc}
		\toprule
		\multirow{2}{*}{Methods} & FLOPs & Params & Training \\ 
        & /sample & size(Mb) & size(Mb) \\
        \midrule
        PointNet \cite{qi2017pointnet}     & 878M & 14 & 60 \\
        PointNet++ \cite{qi2017pointnet++} & 1.69G & 6 & 138 \\
        DGCNN \cite{wang2019dynamic}       & 4.78G & 7 & 185 \\
        KPConv \cite{thomas2019kpconv}     & 200M & - & - \\
        PointCNN \cite{li2018pointcnn}     & 210M & 1 & 22\\
        Pnt Tfmer \cite{engel2021point}    & 92G & 88 & 5736 \\
        Pnt Tfmer \cite{zhao2021point}     & 3.6G & 38 & 6334 \\
        PCT~\cite{guo2020pct}              & 70G & 12 & 2400\\
        DensePoint \cite{liu2019densepoint}& 651M & - & - \\
        RS-CNN \cite{liu2019relation}      & 295M & - & - \\
        \midrule
        SA-CNN (ours) & \textbf{8M} & \textbf{0.2} & \textbf{2}\\
        \bottomrule
	\end{tabular}
	\label{tab:model_efficiency}
\end{table}

\begin{figure}[t]
  \begin{center}
    \includegraphics[width=0.81\linewidth]{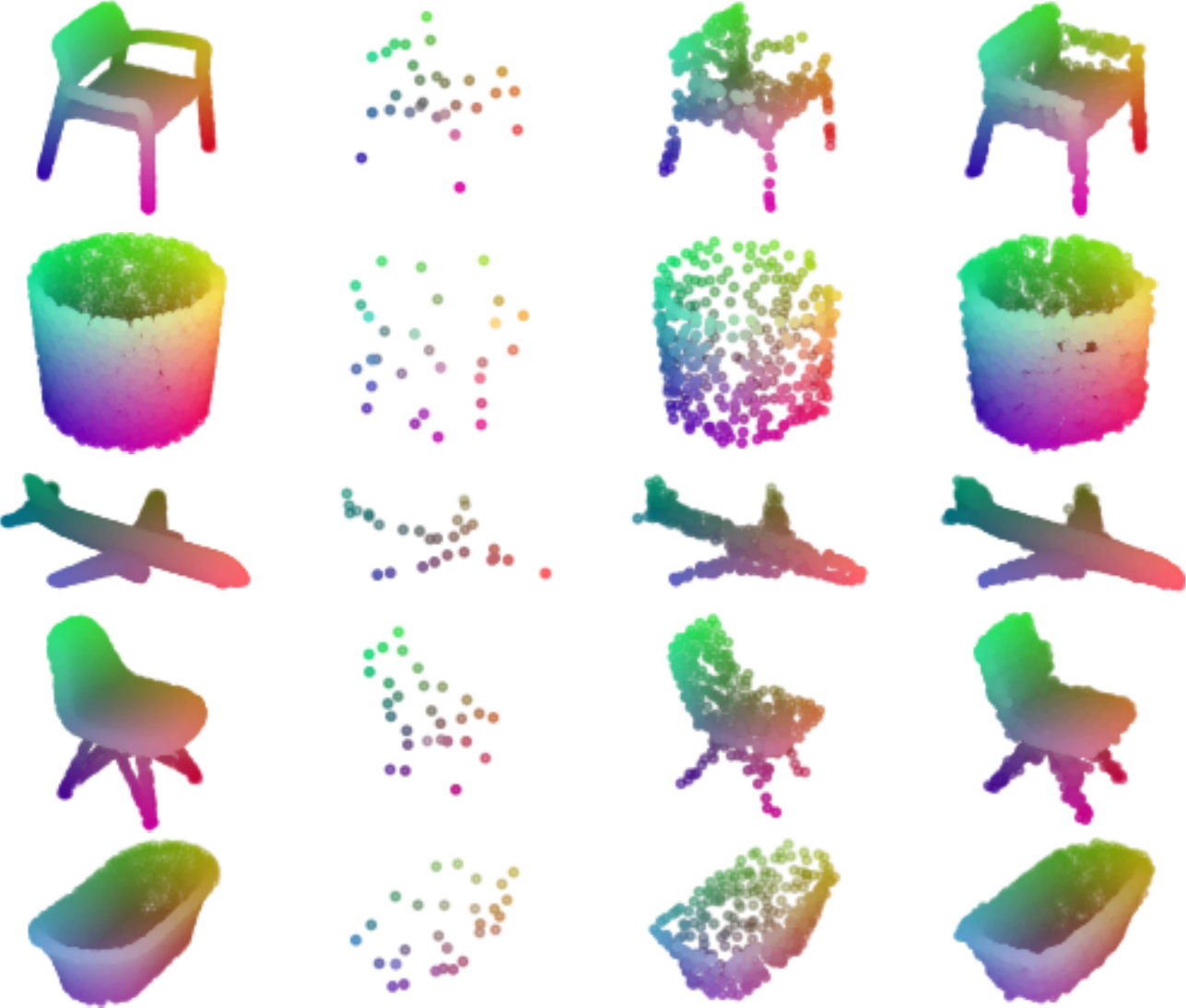}
  \end{center}
  \caption{Hierarchical point cloud decoding on ShapeNet with 2048 points. Left column is the ground truth and right columns are the reconstructions.}
\label{fig:rcs_shapenet2048}
\end{figure}

\begin{figure}[t]
  \begin{center}
    \includegraphics[width=0.81\linewidth]{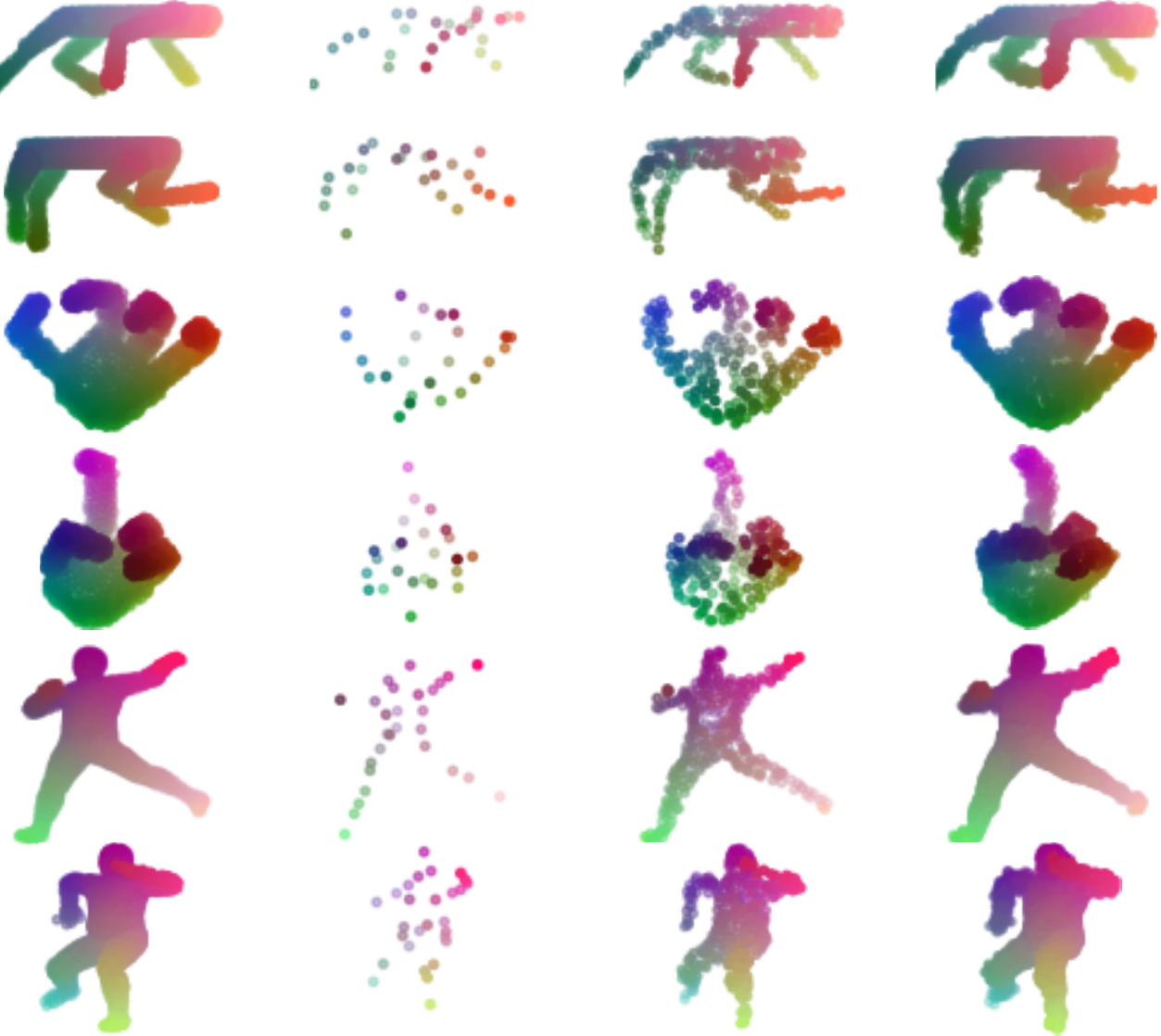}
  \end{center}
  \caption{Hierarchical point cloud decoding on robots and human. Left column is the ground truth and right columns are the reconstructions.}
\label{fig:rcs_smplbot}
\end{figure}

Table \ref{tab:classification} shows the comparison of classification results in term of overall accuracy and the number of parameters on ModelNet40. Statistic in term of floating point operations (FLOPs) per sample and memory for parameter and training required on GPU are depicted in Table \ref{tab:model_efficiency}. 
SA-CNN classifier has 4/5/6/7 as number of head and attention size, 8/7/6/16 as the neighbourhood size, with sampling ratio $0.3N/0.075N/0.015N$, feature size 16/32/64/96 in the 4 levels, and a $p=0.6$ dropout before the output layers.

Due to the simple and efficient hierarchical architecture SA-CNN achieves the lowest number of parameter, FLOPs and memory by several order of magnitude, while having comparable accuracy to the other self-attention and non-attention methods. This advantage makes SA-CNN suitable for low computation/memory point cloud applications. Although generic, the accuracy gap (1.4\%) to the state-of-the-art suggests that SA-CNN could work with additional mechanism \cite{yan2020pointasnl, engel2021point, yang2019modeling} to boost classification performance.

\subsection{Part Segmentation}
\begin{table*}[t]
    \small
    \caption{Part segmentation results (IoU $\%$) on ShapeNet part benchmark. Inputs are either 3D points, or points with normal (n).}
	\centering
	\setlength{\tabcolsep}{2,2pt}
	\begin{tabular}{l c | c | c | c c c c c c c c c c c c c c c c}
		\specialrule{.1em}{.05em}{.05em}
		\multirow{2}{*}{method} & \multirow{2}{*}{input} & cat. & int. & air & \multirow{2}{*}{bag} & \multirow{2}{*}{cap} & \multirow{2}{*}{car} & \multirow{2}{*}{chair} & ear & \multirow{2}{*}{guitar} & \multirow{2}{*}{knife} & \multirow{2}{*}{lamp} & \multirow{2}{*}{laptop} & motor & \multirow{2}{*}{mug} & \multirow{2}{*}{pistol} & \multirow{2}{*}{rocket} & skate & \multirow{2}{*}{table}\\
        &  & mIoU & mIoU & plane &  &  &  &  & phone &  &  &  &  & bike &  &  & &  board & \\
        \specialrule{.1em}{.05em}{.05em}
        SCN\cite{xie2018attentional} & 1k & 81.8 & 84.6 & 83.8 & 80.8 & 83.5 & 79.3 & 90.5 & 69.8 & 91.7 & 86.5 & 82.9 & 96.0 & 69.2 & 93.8 & 82.5 & 62.9 & 74.4 & 80.8\\
        Kd-Net\cite{klokov2017escape} & 4k & 77.4 & 82.3 & 80.1 & 74.6 & 74.3 & 70.3 & 88.6 & 73.5 & 90.2 & 87.2 & 81.0 & 94.9 & 57.4 & 86.7 & 78.1 & 51.8 & 69.9 & 80.3\\
        PointNet\cite{qi2017pointnet} & 2k & 80.4 & 83.7 & 83.4 & 78.7 & 82.5 & 74.9 & 89.6 & 73.0 & 91.5 & 85.9 & 80.8 & 95.3 & 65.2 & 93.0 & 81.2 & 57.9 & 72.8 & 80.6\\
        PointNet++\cite{qi2017pointnet++} & 2k,n & 81.9 & 85.1 & 82.4 & 79.0 & 87.7 & 77.3 & 90.8 & 71.8 & 91.0 & 85.9 & 83.7 & 95.3 & 71.6 & 94.1 & 81.3 & 58.7 & 76.4 & 82.6\\
        PCNN\cite{atzmon2018point} & 2k & 81.8 & 85.1 & 82.4 & 80.1 & 85.5 & 79.5 & 90.8 & 73.2 & 91.3 & 86.0 & 85.0 & 95.7 & 73.2 & 94.8 & 83.3 & 51.0 & 75.0 & 81.8\\
        DGCNN\cite{wang2019dynamic} & 2k & 82.3 & 85.1 & 84.2 & 83.7 & 84.4 & 77.1 & 90.9 & 78.5 & 91.5 & 87.3 & 82.9 & 96.0 & 67.8 & 93.3 & 82.6 & 59.7 & 75.5 & 82.0\\
        RS-CNN\cite{liu2019relation} & 2k & 84.0 & 86.2 & 83.5 & 84.8 & 88.8 & \textbf{79.6} & 91.2 & 81.1 & 91.6 & \textbf{88.4} & 86.0 & 96.0 & 73.7 & 94.1 & 83.4 & 60.5 & 77.7 & 83.6\\
        DensePnt\cite{liu2019densepoint} & 2k & \textbf{84.2} & 86.4 & 84.0 & 85.4 & \textbf{90.0} & 79.2 & 91.1 & 81.6 & 91.5 & 87.5 & 84.7 & 95.9 & 74.3 & 94.6 & 82.9 & \textbf{64.6} & 76.8 & 83.7\\
        Point Tfmer\cite{zhao2021point} & - & 83.7 & 86.6 &-&-&-&-&-&-&-&-&-&-&-&-&-&-&-&- \\
        AttPNet \cite{yang2020attpnet} & 2k & 82.8 & 85.2 &-&-&-&-&-&-&-&-&-&-&-&-&-&-&-&- \\
        GAPNet \cite{chen2019gapnet} & 2k & 82.0 & 84.7 & 84.2 & 84.1 & 88.8 & 78.1 & 90.7 & 70.1 & 91.0 & 87.3 & 83.1 & \textbf{96.2} & 65.9 & 95.0 & 81.7 & 60.7 & 74.9 & 80.8 \\
        LAE-Conv \cite{feng2020point} & 2k & 84.1 & 85.9 & 83.3 & \textbf{86.1} & 85.7 & 80.3 & 90.5 & \textbf{82.7} & 91.5 & 88.1 & 85.5 & 95.9 & \textbf{77.9} & \textbf{95.1} & \textbf{84.0} & 64.3 & 77.6 & 82.8 \\
        \midrule
        SA-CNN (Ours) & 2k & 84.0 & \textbf{86.7} & \textbf{85.1} & 85.3 & 85.4 & 78.6 & \textbf{91.7} & 79.4 & 91.4 & 87.6 & \textbf{86.8} & 95.5 & 72.2 & 94.6 & 81.3 & 62.3 & \textbf{81.5} & \textbf{84.8} \\
        \specialrule{.1em}{.05em}{.05em}
	\end{tabular}
	\label{tab:result_segm}
\end{table*}

\begin{figure}[t]
  \begin{center}
    \includegraphics[width=1\linewidth]{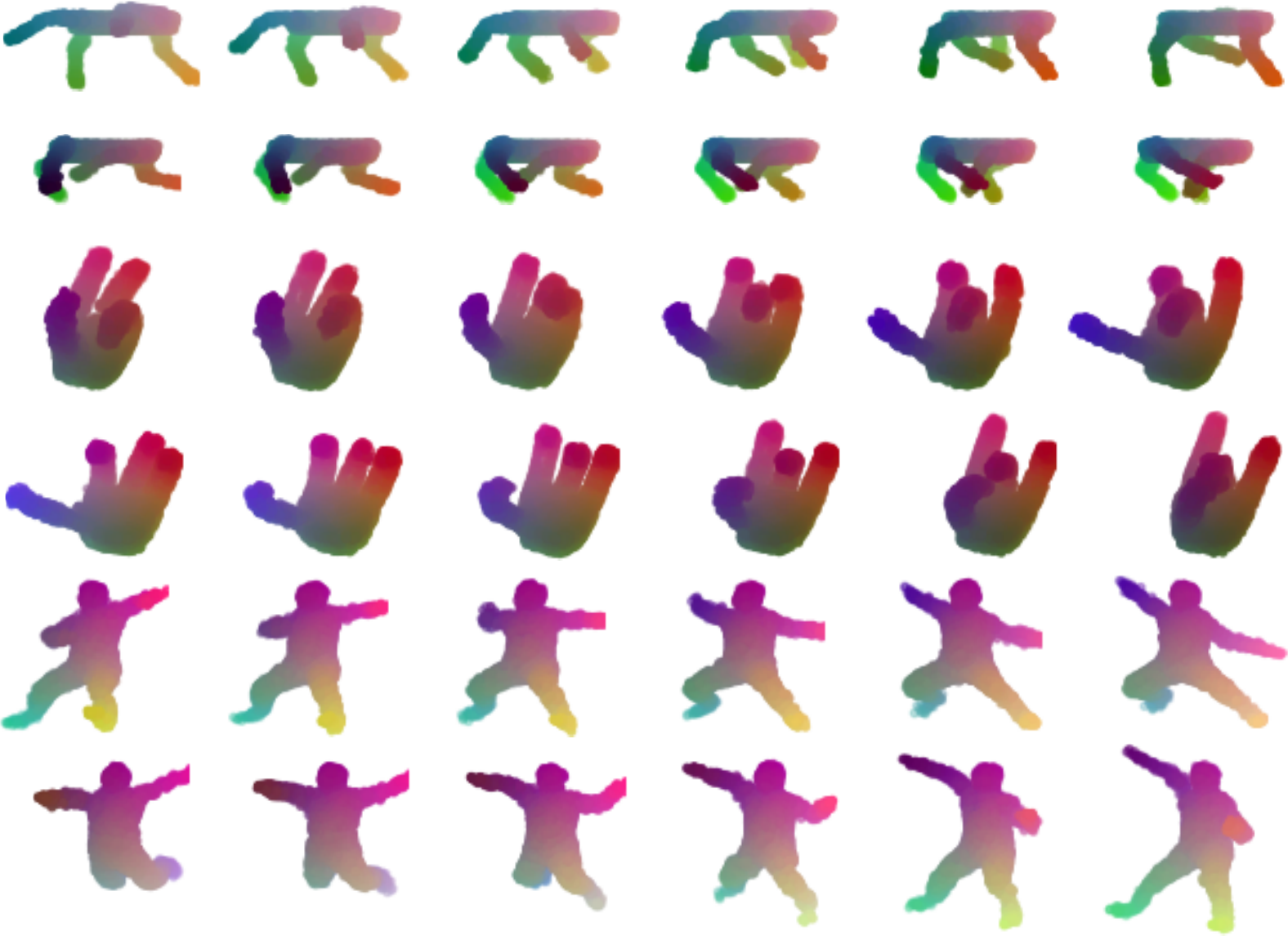}
  \end{center}
  \caption{Latent walk in robots and human models.}
\label{fig:walk_smplbot}
\end{figure}

We evaluate SA-CNN on fine-grained object recognition using ShapeNet part \cite{yi2016scalable} benchmark in point cloud parts segmentation task which consists of 12137 models for training and 2874 for testing. Models are classified into 16 categories, and each category contains between 2 to 6 parts. For each model we zero the bounding box's center and normalize to a unit sphere. During training we augment data with random rotation between [-10, 10] degree, scaling between [0.8, 1.25] and translating between [-0.1, 0.1] in all 3 axes. The performance metric used for this task is intersection-over-union (IoU). There are 3 IoUs reported: IoU for each categories, mean IoU (mIoU) over all categories, and mIoU over all instances. 

\begin{figure}[t]
  \begin{center}
  \includegraphics[width=1\linewidth]{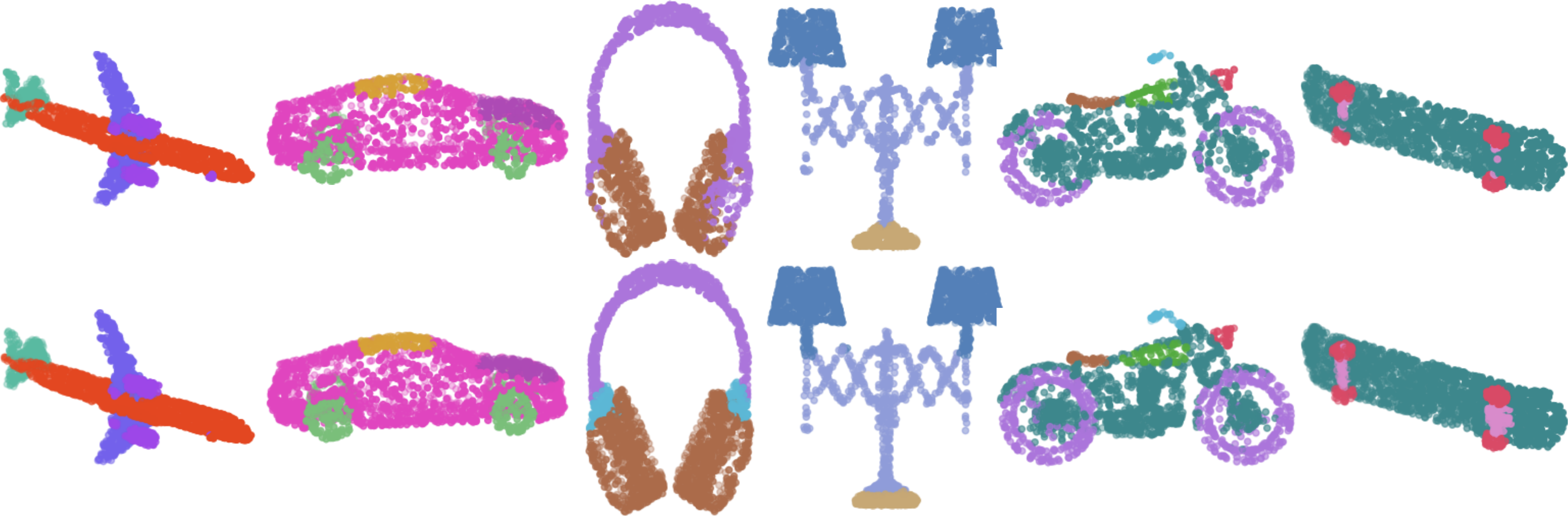}
  \end{center}
  \caption{Examples of ShapeNet part segmentation which contains 16 categories, each having 2 to 6 parts. Top row is segmentation predictions by SA-CNN model, and the bottom row is the corresponding ground truths.}
\label{fig:segmentation}
\end{figure}

Quantitative comparison among the other methods are summarised in Table \ref{tab:result_segm}. SA-CNN achieves state-of-the-art on instance mIoU and several category IoUs. These results demonstrate SA-CNN's capability and robustness in recognising a diverse range of fine-grained shapes. Fig \ref{fig:segmentation} depicts some examples of the ShapeNet segmentation results. 

\subsection{Reconstruction with Auto-Encoder}
In this section, we demonstrate the performance of SA-CNN in auto-encoding. We train the auto-encoder using models from two different domains. For rigid object reconstruction, we use models from ModelNet40 \cite{wu20153d} and ShapeNetCore \cite{chang2015shapenet} with 40 and 55 classes respectively. For non-rigid object reconstruction, we use SMPL \cite{loper2015smpl} human, allegro hand and aliengo quadruped robot model. All non-rigid dataset consists of 4096 synthetically generated training models of different poses. We uniformly sample 2048 points as input as well as output for the training of auto-encoder, and visualize the hierarchical decoding and reconstructions with latent size of 128 in Fig \ref{fig:rcs_shapenet2048} and Fig \ref{fig:rcs_smplbot}. 

\begin{figure}[t]
  \begin{center}
    \includegraphics[width=1\linewidth]{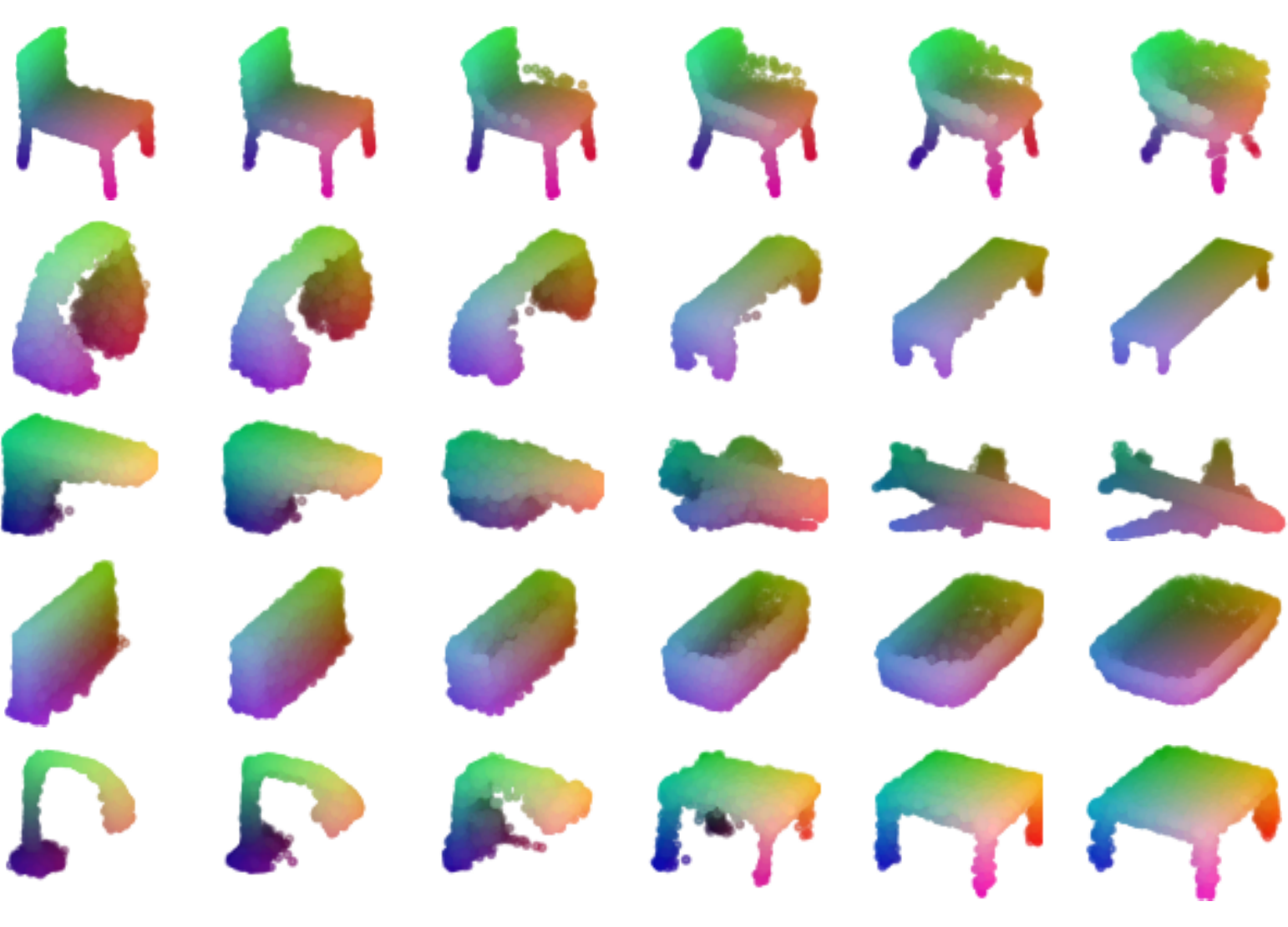}
  \end{center}
  \caption{Latent walk among ShapeNetCore objects.}
\label{fig:walk_shapenet2048}
\end{figure}

Next, we evaluate SA-CNN qualitatively on auto-encoding by visualizing the latent walks of both rigid and non-rigid objects. Linear interpolations of latent code between 2 models are fed to the decoder to get the output reconstructions. Fig \ref{fig:walk_smplbot} and Fig \ref{fig:walk_shapenet2048} depict some examples of latent walks.

\begin{table}[!]
    \small
    \centering
    \caption{The shape retrieval results on the ModelNet40.}
    \vspace{1mm}
	\begin{tabular}{c c | c }
		\toprule
		Methods & Latent Size & mAP (\%) \\
        \midrule
        PointNet\cite{qi2017pointnet} & - & 70.5    \\
        PointCNN\cite{li2018pointcnn} & - & 83.8    \\
        DGCNN\cite{wang2019dynamic}   & - & 85.3    \\
        DensePnt\cite{liu2019densepoint} & 256 & \underline{88.5} \\
        \midrule
        AE & 32 & 84.0 \\
        AE & 64 & 85.4\\
        AE & 128 & 86.7\\
        AE & 256 & 87.1\\
        AE + Triplet & 32 & \underline{88.9}\\
        AE + Triplet & 64 & 89.2\\
        AE + Triplet & 128 & \textbf{90.1}\\
        AE + Triplet & 256 & 89.7\\
        \bottomrule
	\end{tabular}
	\label{tab:shape_retrieval}
\end{table}

\subsection{Shape Retrieval}
We evaluate the performance of SA-CNN auto-encoding with shape retrieval task. We train the auto-encoder using ModelNet40. For this task, given a query shape in test set, we first extract its latent vector using the encoder. Then we search for the top 10 nearest latent vectors seen during training by cosine distance.
We report the mean average precision (mAP) based on the retrieved nearest neighbours. 

\begin{figure}[t]
  \begin{center}
  \includegraphics[width=1\linewidth]{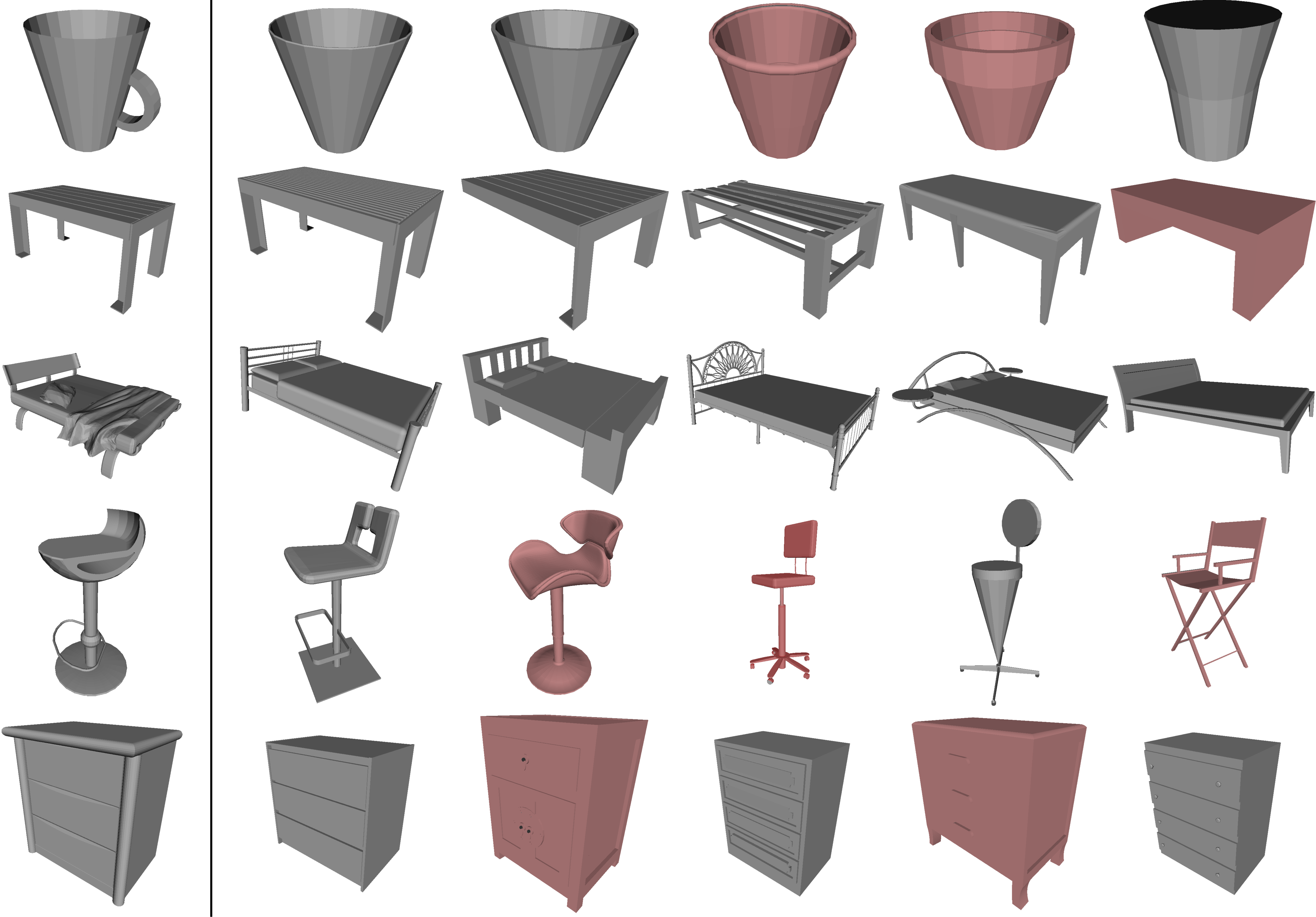}
  \end{center}
  \caption{Examples of ModelNet Shape retrieval task. Given a query model (left) from the test set, top nearest neighbours (right) from the training set are retrieved. Neighbours from the wrong class are highlighted in red.}
\label{fig:retrieval}
\end{figure}

Table \ref{tab:shape_retrieval} summarizes the shape retrieval performance trained with various latent sizes. With pure unsupervised training, our auto-encoder is able to match with methods that train with classification model. With just 32 latent size our method that train with reconstruction and triplet loss surpasses all other methods. Fig \ref{fig:retrieval} depicts some examples of queries and the corresponding nearest neighbours. It is observed that some instances from classes such as cup and pot, and chair and stool, are highly similar, which leads to lower performance in classification by retrieval.

\begin{table}[t]
    \small
    \centering
    \caption{Unsupervised Classification on ModelNet40 using latent representation pre-trained on ShapeNetCore.}
    \vspace{1mm}
	\begin{tabular}{c c | c }
		\toprule
		Methods & Latent Size & Accuracy (\%) \\
        \midrule
        VConv-DAE \cite{sharma2016vconv} & 6912 & 75.5 \\
        3D-GAN \cite{wu2016learning} & 448 & 83.3 \\
        MRTNet-VAE \cite{gadelha2018multiresolution} & 224 & 86.4 \\
        AE-CD \cite{achlioptas2018learning} & 512 & 84.5 \\
        FoldingNet \cite{yang2018foldingnet} & 512 & 88.4 \\
        \midrule
        SA-CNN (Ours) & 128 & 87.1 \\
        SA-CNN (Ours) & 256 & \textbf{88.6} \\
        \bottomrule
	\end{tabular}
	\label{tab:unsupervised_classification}
\end{table}

\subsection{Unsupervised Classification}
In this section, we evaluate the quality of SA-CNN representation learning by comparing classification accuracy using only the latent code as the input feature of a linear Support Vector Machine (SVM). We pre-train the auto-encoder using ShapeNetCore, and then fit and test the SVM using the latent code of ModelNet40 train/test split. 
Table \ref{tab:unsupervised_classification} depicts the latent size and results of the task. Our auto-encoder achieves the state-of-the-art with only 256 latent size show that SA-CNN is capable of compact and efficient unsupervised representation learning for point cloud data.

\subsection{Model Complexity}
Model complexity is an important aspect while using self-attention mechanism. In addition to Table \ref{tab:model_efficiency}, we further summaries model complexity in term of number of parameters and FLOPs per sample of SA-CNN's part segmentation and auto-encoding models in Table \ref{tab:model_complexity}. 
 
SA-CNN part segmentation network has 4/5/6/7/8/8/7/6 number of head and attention size, $0.4N,\,0.12N,\,0.024N$ sampling ratio, 14/12/10/8 neighbourhood size, 16/32/64/96/128/128/96/64 feature dimension in the levels, and uses \textrm{Max} pooling for the first 3 and \textrm{Avg} pooling for the rest of the SAConv. Auto-encoding networks have similar settings but increase proportionally to the latent size.
 
\begin{table}[!]
    \small
    \centering
    \caption{SA-CNN's part segmentation and auto-encoding model complexity in number of parameter and FLOPs per sample.}
    \vspace{1mm}
	\begin{tabular}{cccc}
		\toprule
		Network & Setting & \#param & \#FLOPs\\
        \midrule
        \multirow{2}{*}{Part Segmentation} & 1k points & 227k & 1.2G \\
         & 2k points & 227k & 2.4G \\
        \midrule
        \multirow{5}{*}{Encoder/Decoder} & 32 latent & 22k/35k & 15M/20M\\
        & 64 latent & 58k/77k & 38M/80M\\
        & 128 latent & 160k/190k & 100M/180M\\
        & 128 latent & \multirow{2}{*}{160k/190k} & \multirow{2}{*}{116M/271M}\\
        & 2k points & & \\
        \bottomrule
	\end{tabular}
	\label{tab:model_complexity}
\end{table}

\section{Conclusion}

In this work, we present SA-CNN, a lightweight self-attention based encoding and decoding architecture for 3D point cloud representation learning. The proposed SA-CNN is a drop-in replacement for multiple hierarchical point cloud analysis tasks. We demonstrate that SA-CNN achieves the state-of-the-art or comparable performance to other methods on multiple benchmarks with significantly lower model complexity. In addition, we visualize the multi-stages point cloud reconstruction and their latent walks on rigid objects as well as non-rigid human and robot models.

{\small
\bibliographystyle{IEEEtran}
\bibliography{main}
}
\end{document}